\documentclass[11pt,a4paper]{article}
\usepackage[hyperref]{acl2019}
\usepackage{times}
\usepackage{latexsym}
\usepackage{url}
\usepackage{graphicx}

\aclfinalcopy 

\title{Whatcha lookin' at? DeepLIFTing BERT's Attention in Question Answering}

\author{Ekaterina Arkhangelskaia \\
	Saarland University \\
	\texttt{s8ekarkh} \\
	\texttt{@stud.uni-saarland.de}
	\\\And
	Sourav Dutta \\
	Saarland University \\
	\texttt{s8sodutt} \\
	\texttt{@stud.uni-saarland.de}}

\date{}

\begin{document}
\maketitle
\begin{abstract}
	There has been great success recently in tackling challenging NLP tasks by neural networks
	which have been pre-trained and fine-tuned on large amounts of task data. In this pa-
	per, we investigate one such model, BERT for question-answering, with the aim to analyze
	why it is able to achieve significantly better results than other models. We run DeepLIFT
	on the model predictions and test the outcomes to monitor shift in the attention values for input. We also cluster the results to analyze any possible patterns similar to human reasoning depending on the kind of input paragraph and question the model is trying to answer.
\end{abstract}

\section{Introduction}
In the last couple of years, neural network models trained on large text data and fine-tuned on supervised tasks, have been rapidly advancing the state-of-the-art benchmarks in Natural Language Processing (NLP). Recent models like ELMo \cite{peters2018deep}, GPT \cite{radford2018improving}, and BERT \cite{devlin2019bert} are gradually replacing the word embedding models like Word2Vec \cite{mikolov2013distributed} and GloVe \cite{pennington2014glove} as the goto approaches for tackling NLP tasks. The recent success in NLP tasks apparently mean that neural networks must be able to learn syntactic, semantic, and/or certain other linguistic information from input training data. However due to their blackbox nature, no one knows exactly why neural networks are able to outperform previous state-of-the-art methods by such a big margin.

Recently there is a growing interest in solving this mystery of how and why neural networks
work the way they do. Previously, while some researchers have tried to observe the internal hidden vector representations of models by applying methods like probing classifiers \cite{belinkov2017neural}, others have examined the outputs of language models by varying the input data \cite{linzen2016assessing}. There also has been recent research on analyzing how attention works in such models \cite{DBLP:journals/corr/abs-1906-04341}. All these works have produced evidence that deep neural language models are capable of encoding some form of syntactic and semantic information. This linguistic knowledge enables models like BERT to tackle challenging tasks in the classical NLP pipeline \cite{tenney2019bert}. One of the main components behind the recent massive success of neural networks, especially in NLP tasks, is attention. Attention \cite{bahdanau2014neural} is simply the parameter that determines how important the past data is, given the current context. It is a weight matrix that helps in calculating the next representation for the current word in text.

Like \citet{DBLP:journals/corr/abs-1906-04341}, we here analyze all the attention heads of BERT, except that we inspect a BERT model pre-trained for question-answering on the SQuaD 2.0 dataset \cite{rajpurkar2018know} with the aim to find how important are different parts of the input on each attention layer. We first extract the attention values of each layer in the forward pass during model training and run DeepLIFT \cite{shrikumar2017learning} on the results to determine the contribution of each attention component on the output for a given input question. We then try to detect patterns of shifting attentions by clustering the resulting representations and analyzing questions typical for each cluster.

\noindent \textbf{Outline of the report.} Having briefly discussed some of the background terminologies in section 2, we mention some related work in section 3. Section 4 explains our approach and we analyze the results of our experiment in section 5. We conclude in section 6.

\section{Background}

\subsection{BERT for Question Answering}
BERT \cite{devlin2019bert} is a large neural network model based on the transformer architecture that is pre-trained on task specific data. Transformers \cite{vaswani2017attention} are large networks made up of multiple encoder-decoder layers, each layer containing multi-headed attention.

BERT (Bidirectional Encoder Representations
from Transformers) is primarily trained for two different tasks; masked language modeling where the model tries to predict words that have been removed or masked, and next sentence prediction where it tries to guess whether a statement follows a given proposition or not. BERT is pre-trained on 3.3 billion English text tokens and then fine-tuned on supervised task specific domain data to produce impressive results. Special tokens [CLS] and [SEP] are added to the beginning and end of the text respectively. We here use the base version of BERT which has 12 transformer layers containing 12 attention heads each, thus a total of 144 attention heads.

SQuaD \cite{rajpurkar2016squad} is a dataset containing a list of questions and answers. Our model is fine-tuned on the updated SQuaD 2.0 dataset \cite{rajpurkar2018know} which also tells us if a particular question is answerable given the input paragraph context.

\subsection{DeepLIFT}
Researchers have used different gradient-based attribution methods to analyze the flow of information inside Deep Neural Networks (DNNs). There are perturbation-based methods like Occlusion \cite{zeiler2014visualizing} where output change is monitored on replacing a single feature with a zero baseline. Similarly there are methods replying on backpropagation like Gradient*Input (Shrikumar et al., 2016), Integrated Gradients \cite{sundararajan2017axiomatic}, and Layer-wise Relevance Propagation (LRP) \cite{bach2015pixel}. We here use DeepLIFT (Deep Learning Important FeaTures) \cite{shrikumar2017learning} for this purpose. According to the authors, DeepLIFT is
a method to decompose the prediction of a neural network for a specific input by backpropagating once layer-wise through the model architecture and monitoring the contribution of each neuron to every input feature. DeepLIFT assigns separate values for positive and negative contributions, and thus is able to reveal dependencies that other methods might miss. It also avoids placing misleading importance on bias terms. A comparative
case-study of different attribution methods \cite{ancona2018towards} shows that DeepLIFT has high correlation and it is a faster and better approximation of Integrated Gradients, making it a good choice for our analysis.

We run DeepLIFT on the final results of our model, using the highest probability start and end
words as the target neurons. DeepLIFT, generally defined for feed-forward networks only, gets
more complicated for multi-headed multi-layer attention with multiple inputs and inner products. We had to rewrite the backward pass of DeepLIFT ourselves as PyTorch\footnote{\url{https://pytorch.org/}} does not support backward hooks for complex modules, like the ones utilized by BERT. We also had to change the propagation algorithm from the original DeepLIFT paper, since computing multipliers in the forward pass takes up more memory quadratically, compared to our implementation.

\section{Related Work}
As mentioned before, researchers have tried to unravel the mystery of why neural networks work so well. There have been some recent work on analyzing the BERT model to understand if it is able to encode and learn linguistic information from given input text data.

\citet{DBLP:journals/corr/abs-1906-04341} analyzed what BERT looks at and found evidence that attention heads in BERT attend to patterns in data like the next token, delimiters, and periods, with the same layer often exhibiting similar behavior. Certain attention heads
can relate to specific linguistic information like syntax and coreference. Substantial amount of language representations can be found in BERT’s attention maps. Researchers have also found proof that BERT is able to represent the classical NLP pipeline \cite{tenney2019bert} in a localized interpretable manner in the sequence of POS tagging, parsing, NER, semantics, and coreference. It is further proved \cite{jawahar:hal-02131630} that BERT captures phrasal information in its lower layers, followed by syntactic and finally semantic representations as it goes through the upper layers. Deep neural networks with higher number of layers are better suited to capture long-distance dependency information from input text data.

\section{Method}
Here we use a BERT model fine-tuned on the SQuaD 2.0 dataset. We are able to obtain high accuracy scores, comparable to state of the art benchmark results. However, our main objective
in this experiment is not to come up with a new model that would beat the current state of the
art. Rather, we want to investigate how the multi-headed attention mechanism works in BERT and
how similar are the changes in focus on different input tokens to human thought processes.

\begin{center}
\begin{verbatim}
 [CLS]question[SEP]paragraph[SEP]
\end{verbatim}
\end{center}

The test data is fed into the model in this format. We want to monitor changes in the amount of attention that is given to each of the tokens in input text. We run DeepLIFT on the results by backpropagating through the layers of our neural network model. DeepLIFT produces certain scores for each token in the input text for each layer. The scores are either positive or negative, representing higher or lower attention on the tokens respectively. We want to focus on those units which receive higher attention in the process, and the shift
in those values. The tokens are highlighted with colors which represent their DeepLIFT scores.

Here we present an example from our experiment. The question is \emph{when did beyonce start becoming popular?}, to which the answer is \emph{late 1990s}.

The code for this experiment will be open-sourced.

\section{Results and Analysis}
Please refer to the images included in the Supplemental Material section of this report relevant for both the examples mentioned above.

\noindent \textbf{Example 1. \emph{when did beyonce start becoming popular?}} Figures 1 to 12 show us which tokens are given more attention by our model. The blue color represents low attention while red indicates higher values (check the reference scale provided with each image). In the initial layers, the model focuses on the separator tokens in the input text
first. Then it switches the focus to punctuation symbols and gradually to certain tokens in text. This behavior relates to BERT focusing on the syntactic information of input \cite{DBLP:journals/corr/abs-1906-04341} in the initial layers. On the other hand, the model shifts its attention to tokens which are most likely related to the question and can be important part of the probable answer. This captures the semantic representations of the text. We can see how attention changes for a token if we look at the word \emph{beyonce} in figures 4 and 5. The model focus on \emph{beyonce} till layer 4 and then completely ignores (no attention) it from layer 5. It was focusing on the token as it is a keyword in the question. It removes its focus when it tries to find its answer in the text. It focuses on the question keywords (for example \emph{beyonce}) again in the last layers to semantically
verify the answer context with the question keywords. In the output layer (figure 13), we see our model gives its full attention to only those tokens which are part of the prediction (answer).

\section{Conclusion}
Here we have used DeepLIFT, a backpropagation-based attribution method, to first analyze how the values of multi-headed attention change across the layers of BERT and then clustered the results to find patterns in the data similar to human reasoning. We find that BERT, fine-tuned for question-answering tasks, first focuses on the tokens in the text with respect to keywords in the question. Later on it shifts its attention to only those tokens which it thinks are vital in constructing the final answer to the question. We pictorially demonstrate how the model’s widespread span of attention narrows down to the answer tokens in the final layers, keeping in mind the syntactic and semantic representations during the entire process.

\section*{Acknowledgments}

This report is part of the \emph{Neural Question Answering} seminar offered in Saarland University, Germany. We are grateful to Prof. Günter Neumann and Stalin Varanasi for offering this seminar and guiding us.

\bibliography{acl2019}

\begin{thebibliography}{20}
\expandafter\ifx\csname natexlab\endcsname\relax\def\natexlab#1{#1}\fi

\bibitem[{Ancona et~al.(2018)Ancona, Ceolini, Oztireli, and
  Gross}]{ancona2018towards}
Marco Ancona, Enea Ceolini, Cengiz Oztireli, and Markus Gross. 2018.
\newblock Towards better understanding of gradient-based attribution methods
  for deep neural networks.
\newblock In \emph{6th International Conference on Learning Representations
  (ICLR 2018)}.

\bibitem[{Bach et~al.(2015)Bach, Binder, Montavon, Klauschen, M{\"u}ller, and
  Samek}]{bach2015pixel}
Sebastian Bach, Alexander Binder, Gr{\'e}goire Montavon, Frederick Klauschen,
  Klaus-Robert M{\"u}ller, and Wojciech Samek. 2015.
\newblock On pixel-wise explanations for non-linear classifier decisions by
  layer-wise relevance propagation.
\newblock \emph{PloS one}, 10(7):e0130140.

\bibitem[{Bahdanau et~al.(2014)Bahdanau, Cho, and Bengio}]{bahdanau2014neural}
Dzmitry Bahdanau, Kyunghyun Cho, and Yoshua Bengio. 2014.
\newblock Neural machine translation by jointly learning to align and
  translate.
\newblock \emph{arXiv preprint arXiv:1409.0473}.

\bibitem[{Belinkov et~al.(2017)Belinkov, Durrani, Dalvi, Sajjad, and
  Glass}]{belinkov2017neural}
Yonatan Belinkov, Nadir Durrani, Fahim Dalvi, Hassan Sajjad, and James Glass.
  2017.
\newblock What do neural machine translation models learn about morphology?
\newblock In \emph{Proceedings of the 55th Annual Meeting of the Association
  for Computational Linguistics (Volume 1: Long Papers)}, pages 861--872.

\bibitem[{Clark et~al.(2019)Clark, Khandelwal, Levy, and
  Manning}]{DBLP:journals/corr/abs-1906-04341}
Kevin Clark, Urvashi Khandelwal, Omer Levy, and Christopher~D. Manning. 2019.
\newblock \href {http://arxiv.org/abs/1906.04341} {What does {BERT} look at? an
  analysis of bert's attention}.
\newblock \emph{CoRR}, abs/1906.04341.

\bibitem[{Devlin et~al.(2019)Devlin, Chang, Lee, and
  Toutanova}]{devlin2019bert}
Jacob Devlin, Ming-Wei Chang, Kenton Lee, and Kristina Toutanova. 2019.
\newblock Bert: Pre-training of deep bidirectional transformers for language
  understanding.
\newblock In \emph{Proceedings of the 2019 Conference of the North American
  Chapter of the Association for Computational Linguistics: Human Language
  Technologies, Volume 1 (Long and Short Papers)}, pages 4171--4186.

\bibitem[{Jawahar et~al.(2019)Jawahar, Sagot, and
  Seddah}]{jawahar:hal-02131630}
Ganesh Jawahar, Beno{\^i}t Sagot, and Djam{\'e} Seddah. 2019.
\newblock \href {https://hal.inria.fr/hal-02131630} {{What does BERT learn
  about the structure of language?}}
\newblock In \emph{{ACL 2019 - 57th Annual Meeting of the Association for
  Computational Linguistics}}, Florence, Italy.

\bibitem[{Linzen et~al.(2016)Linzen, Dupoux, and
  Goldberg}]{linzen2016assessing}
Tal Linzen, Emmanuel Dupoux, and Yoav Goldberg. 2016.
\newblock Assessing the ability of lstms to learn syntax-sensitive
  dependencies.
\newblock \emph{Transactions of the Association for Computational Linguistics},
  4:521--535.

\bibitem[{Mikolov et~al.(2013)Mikolov, Sutskever, Chen, Corrado, and
  Dean}]{mikolov2013distributed}
Tomas Mikolov, Ilya Sutskever, Kai Chen, Greg~S Corrado, and Jeff Dean. 2013.
\newblock Distributed representations of words and phrases and their
  compositionality.
\newblock In \emph{Advances in neural information processing systems}, pages
  3111--3119.

\bibitem[{Pennington et~al.(2014)Pennington, Socher, and
  Manning}]{pennington2014glove}
Jeffrey Pennington, Richard Socher, and Christopher Manning. 2014.
\newblock Glove: Global vectors for word representation.
\newblock In \emph{Proceedings of the 2014 conference on empirical methods in
  natural language processing (EMNLP)}, pages 1532--1543.

\bibitem[{Peters et~al.(2018)Peters, Neumann, Iyyer, Gardner, Clark, Lee, and
  Zettlemoyer}]{peters2018deep}
Matthew~E Peters, Mark Neumann, Mohit Iyyer, Matt Gardner, Christopher Clark,
  Kenton Lee, and Luke Zettlemoyer. 2018.
\newblock Deep contextualized word representations.
\newblock \emph{arXiv preprint arXiv:1802.05365}.

\bibitem[{Radford et~al.(2018)Radford, Narasimhan, Salimans, and
  Sutskever}]{radford2018improving}
Alec Radford, Karthik Narasimhan, Tim Salimans, and Ilya Sutskever. 2018.
\newblock Improving language understanding by generative pre-training.
\newblock \emph{URL https://blog.openai.com/language-unsupervised}.

\bibitem[{Rajpurkar et~al.(2018)Rajpurkar, Jia, and Liang}]{rajpurkar2018know}
Pranav Rajpurkar, Robin Jia, and Percy Liang. 2018.
\newblock Know what you don't know: Unanswerable questions for squad.
\newblock \emph{arXiv preprint arXiv:1806.03822}.

\bibitem[{Rajpurkar et~al.(2016)Rajpurkar, Zhang, Lopyrev, and
  Liang}]{rajpurkar2016squad}
Pranav Rajpurkar, Jian Zhang, Konstantin Lopyrev, and Percy Liang. 2016.
\newblock Squad: 100,000+ questions for machine comprehension of text.
\newblock \emph{arXiv preprint arXiv:1606.05250}.

\bibitem[{Shrikumar et~al.(2017)Shrikumar, Greenside, and
  Kundaje}]{shrikumar2017learning}
Avanti Shrikumar, Peyton Greenside, and Anshul Kundaje. 2017.
\newblock Learning important features through propagating activation
  differences.
\newblock In \emph{Proceedings of the 34th International Conference on Machine
  Learning-Volume 70}, pages 3145--3153. JMLR. org.

\bibitem[{Shrikumar et~al.(2016)Shrikumar, Greenside, Shcherbina, and
  Kundaje}]{shrikumar2016not}
Avanti Shrikumar, Peyton Greenside, Anna Shcherbina, and Anshul Kundaje. 2016.
\newblock Not just a black box: Learning important features through propagating
  activation differences.
\newblock \emph{arXiv preprint arXiv:1605.01713}.

\bibitem[{Sundararajan et~al.(2017)Sundararajan, Taly, and
  Yan}]{sundararajan2017axiomatic}
Mukund Sundararajan, Ankur Taly, and Qiqi Yan. 2017.
\newblock Axiomatic attribution for deep networks.
\newblock In \emph{Proceedings of the 34th International Conference on Machine
  Learning-Volume 70}, pages 3319--3328. JMLR. org.

\bibitem[{Tenney et~al.(2019)Tenney, Das, and Pavlick}]{tenney2019bert}
Ian Tenney, Dipanjan Das, and Ellie Pavlick. 2019.
\newblock Bert rediscovers the classical nlp pipeline.
\newblock \emph{arXiv preprint arXiv:1905.05950}.

\bibitem[{Vaswani et~al.(2017)Vaswani, Shazeer, Parmar, Uszkoreit, Jones,
  Gomez, Kaiser, and Polosukhin}]{vaswani2017attention}
Ashish Vaswani, Noam Shazeer, Niki Parmar, Jakob Uszkoreit, Llion Jones,
  Aidan~N Gomez, {\L}ukasz Kaiser, and Illia Polosukhin. 2017.
\newblock Attention is all you need.
\newblock In \emph{Advances in neural information processing systems}, pages
  5998--6008.

\bibitem[{Zeiler and Fergus(2014)}]{zeiler2014visualizing}
Matthew~D Zeiler and Rob Fergus. 2014.
\newblock Visualizing and understanding convolutional networks.
\newblock In \emph{European conference on computer vision}, pages 818--833.
  Springer.

\end{thebibliography}
\bibliographystyle{acl_natbib}

\appendix


\section{Supplemental Material}
We have included all the attention \emph{heatmap} images on input questions and the corresponding input paragraph for each layer of BERT and its prediction. In each image, tokens of the text data input is highlighted with a specific color to denote the score that is assigned to them after applying DeepLIFT. That score tells us how much attention the model gives on that particular token in the text.

\subsection{Example 1}
\textbf{Q}: \emph{when did beyonce start becoming popular?}

\noindent \textbf{A}: \emph{late 1990s}

\begin{itemize}
	\item \textbf{Figure 1 - 12} show the importance given to each token after running DeepLIFT. 
	\item \textbf{Figure 13} shows amount of attention given by BERT on the tokens in the output. We see the tokens \emph{"late 1990s"} receives highest attention, that being the correct answer.
\end{itemize}

The \textcolor{blue}{blue} highlighted color refers to a negative score, which means that this value contributed negatively to the final result. Similarly, the color \textcolor{red}{red} represents a positive score, that is, more positive contribution to the result. A color scale is provided for reference.

\label{sec:supplemental}
\begin{figure*}
	\centering
	\includegraphics[width=\textwidth]{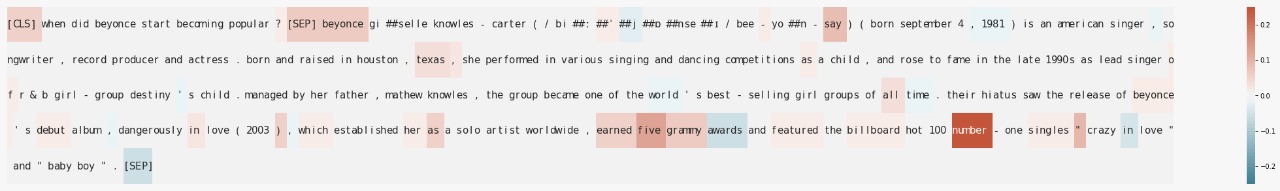}
	\caption{Attention in BERT layer 1}
	\label{Layer 1}
\end{figure*}
\begin{figure*}
	\centering
	\includegraphics[width=\textwidth]{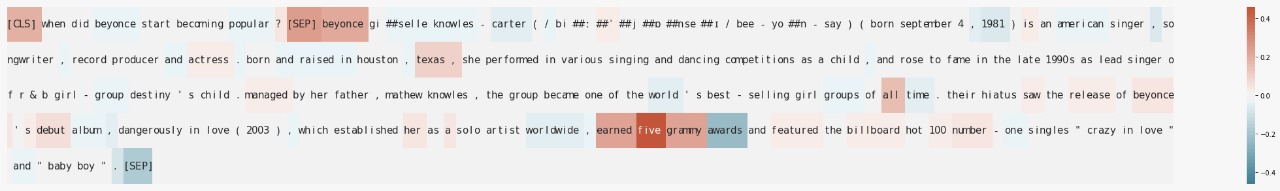}
	\caption{Attention in BERT layer 2}
	\label{Layer 2}
\end{figure*}
\begin{figure*}
	\centering
	\includegraphics[width=\textwidth]{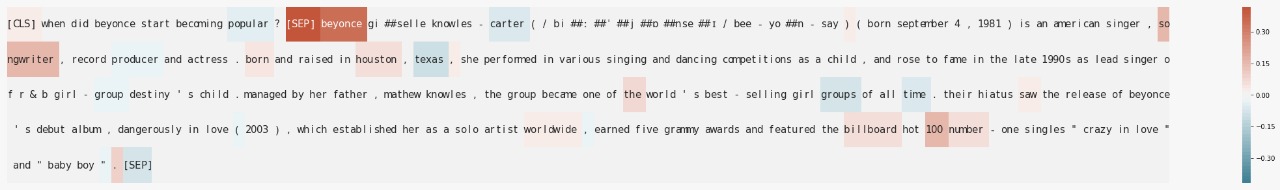}
	\caption{Attention in BERT layer 3}
	\label{Layer 3}
\end{figure*}
\begin{figure*}
	\centering
	\includegraphics[width=\textwidth]{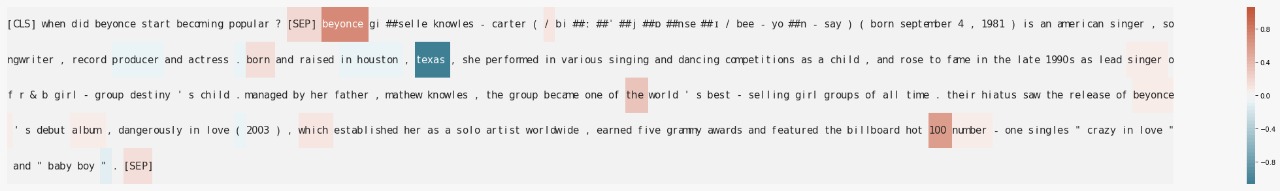}
	\caption{Attention in BERT layer 4}
	\label{Layer 4}
\end{figure*}
\begin{figure*}
	\centering
	\includegraphics[width=\textwidth]{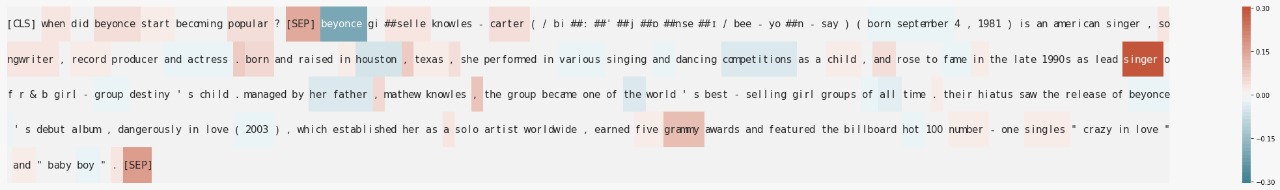}
	\caption{Attention in BERT layer 5}
	\label{Layer 5}
\end{figure*}
\begin{figure*}
	\centering
	\includegraphics[width=\textwidth]{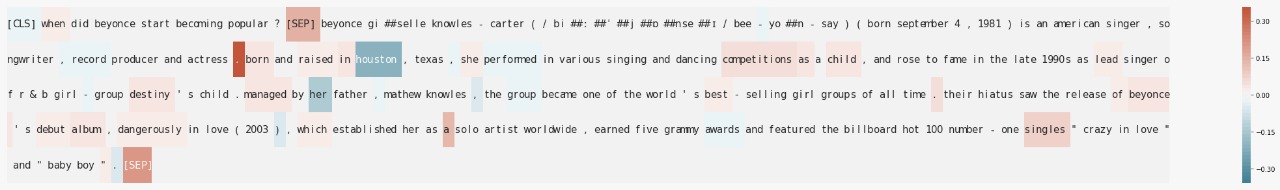}
	\caption{Attention in BERT layer 6}
	\label{Layer 6}
\end{figure*}
\begin{figure*}
	\centering
	\includegraphics[width=\textwidth]{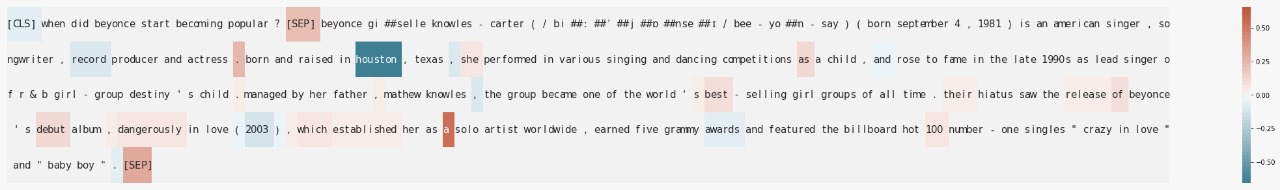}
	\caption{Attention in BERT layer 7}
	\label{Layer 7}
\end{figure*}
\begin{figure*}
	\centering
	\includegraphics[width=\textwidth]{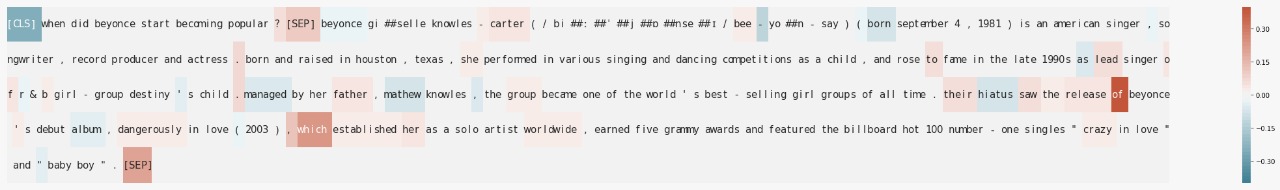}
	\caption{Attention in BERT layer 8}
	\label{Layer 8}
\end{figure*}
\begin{figure*}
	\centering
	\includegraphics[width=\textwidth]{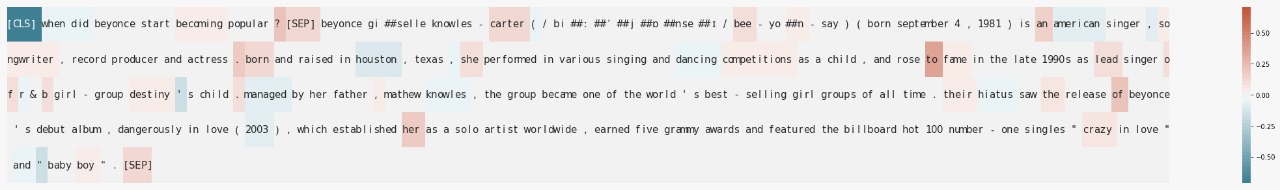}
	\caption{Attention in BERT layer 9}
	\label{Layer 9}
\end{figure*}
\begin{figure*}
	\centering
	\includegraphics[width=\textwidth]{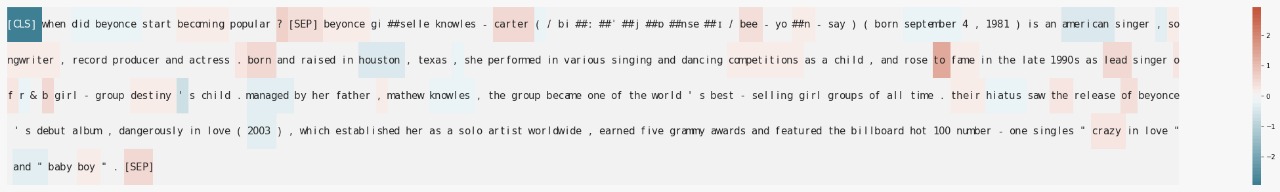}
	\caption{Attention in BERT layer 10}
	\label{Layer 10}
\end{figure*}
\begin{figure*}
	\centering
	\includegraphics[width=\textwidth]{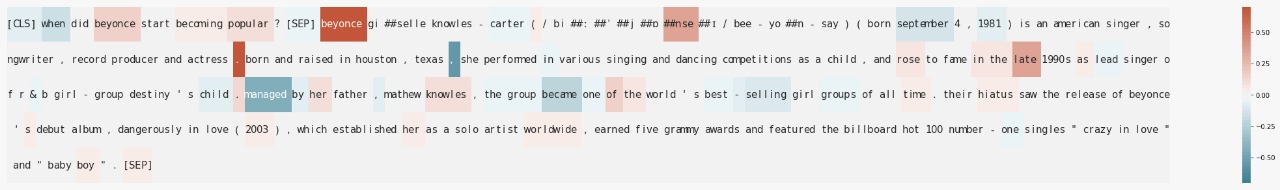}
	\caption{Attention in BERT layer 11}
	\label{Layer 11}
\end{figure*}
\begin{figure*}
	\centering
	\includegraphics[width=\textwidth]{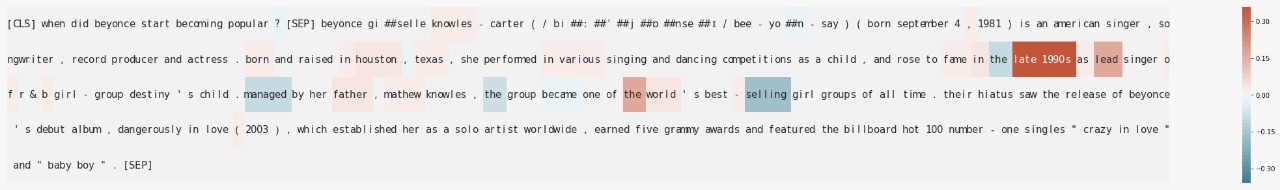}
	\caption{Attention in BERT layer 12}
	\label{Layer 12}
\end{figure*}
\begin{figure*}
	\centering
	\includegraphics[width=\textwidth]{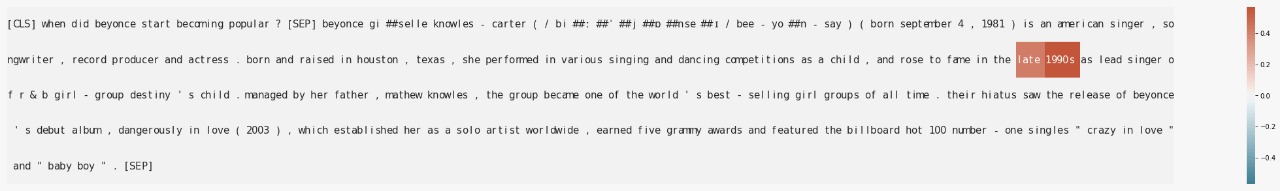}
	\caption{Attention in BERT output}
	\label{output layer}
\end{figure*}

\end{document}